\documentclass[10pt,journal,compsoc]{IEEEtran}

\usepackage{amsmath,amsfonts,amssymb}
\usepackage[scaled=1.05]{newtxtext}
\usepackage{newtxmath}
\usepackage{array}
\usepackage[caption=false,font=footnotesize,labelfont=sf,textfont=sf]{subfig}
\usepackage{textcomp}
\usepackage{url}
\usepackage{graphicx}
\usepackage{cite}
\usepackage{booktabs}
\usepackage{multirow}
\usepackage{threeparttable}
\usepackage{placeins}
\usepackage[table]{xcolor}
\begin{document}

\bstctlcite{IEEEtranBSTCTL:etal}

\title{EgoTSR++: Egocentric Spatiotemporal Reasoning for Task Progress Understanding}

\author{Xiaoda Yang, Can Wang, Yuxiang Liu, Pengfei Zhou, Jianwen Lou, and Shuicheng Yan%
\thanks{Xiaoda Yang and Can Wang contributed equally to this work. Shuicheng Yan is the corresponding author.}
\thanks{Xiaoda Yang and Jianwen Lou are with Zhejiang University, Hangzhou, China. Can Wang is with Qingdao University, Qingdao, China. Yuxiang Liu is with Tianjin University, Tianjin, China. Pengfei Zhou and Shuicheng Yan are with the National University of Singapore, Singapore (e-mail: yansc@nus.edu.sg).}
}

\markboth{EgoTSR++: Egocentric Spatiotemporal Reasoning --- Preprint}{Yang et al.: EgoTSR++}

\IEEEtitleabstractindextext{%
\begin{abstract}
Vision-Language Models (VLMs) have advanced rapidly in static visual understanding, yet remain unreliable when judging how an egocentric task is progressing. Given a task instruction and two visual observations, a model should determine which state is closer to the goal by analyzing task-relevant object configurations and spatial relations, rather than relying on timestamps or presentation order. This distinction is critical in manipulation, where retries, corrective actions, and temporary regressions make progress inherently non-monotonic. We introduce EgoTSR, a unified framework for diagnosing and improving order-robust task-progress understanding. First, SpatialLogic-Bench evaluates each physical state pair in both original and order-swapped presentations across short- and long-horizon settings, exposing whether a model follows task-state evidence or chronological shortcuts. Second, our data construction pipeline converts successful, approximately monotonic manipulation and first-person trajectories into bidirectional supervision; LongTag further preserves intermediate subtask structure for long-horizon comparison, while failure-aware data extend learning to regressions and recoveries. Third, a progressive CoT$\rightarrow$Tag curriculum first supervises evidence-grounded interpretation of task-relevant state changes and then consolidates the comparison rule through scalable label-only training. Experiments reveal substantial input-order bias in representative VLMs. EgoTSR achieves $92.4\%$ long-horizon accuracy with a $0.1$-point forward--inverse Gap. Failure-aware supervision further improves accuracy on non-monotonic trajectories by $11.8$ points and Recovery Accuracy by $11.2$ points, while maintaining broad visual and spatial capabilities. These results establish goal-conditioned state comparison as an explicit formulation of egocentric spatiotemporal reasoning for task-progress understanding.
\end{abstract}

}

\maketitle
\IEEEdisplaynontitleabstractindextext
\IEEEpeerreviewmaketitle

\section{Introduction}

\IEEEPARstart{V}{ision}-Language Models (VLMs) have advanced rapidly in static scene understanding and multimodal semantic modeling~\cite{zhang2024vlmsurvey}. Egocentric video captures activities from an agent-centered view~\cite{grauman2022ego4d}, where objects, hands, tools, and spatial relations evolve over time. Understanding task progress requires connecting these visible state changes to the intended goal. Recent work has diagnosed order sensitivity in task-oriented spatiotemporal reasoning~\cite{yang2026spatiallogic}, extended the capability through curriculum learning~\cite{yang2026perceptionplanning}, and studied progressive supervision for counteracting spatiotemporal hallucinations~\cite{yang2026progressive}.

We make this broad capability measurable through goal-conditioned task-progress comparison. Given a task instruction and two observations from the same execution, the model identifies which state is closer to completion from task-relevant object configurations and spatial relations. Robotic manipulation provides the primary testbed because its progress is expressed through concrete changes in contact, placement, orientation, and object relations. The resulting formulation targets state assessment rather than policy execution itself, while providing a common interface for robotic and human first-person trajectories.

Current VLMs can conflate task progress with chronology or presentation position. SpatialLogic-Bench diagnoses this sensitivity by presenting the same physical pair in both orders~\cite{yang2026spatiallogic}. We call this behavior \emph{input-order bias}: a model's physical preference changes when only the presentation order of the two states is reversed. This is a task-specific instance of shortcut learning~\cite{geirhos2020shortcut}. The distinction is especially consequential in manipulation, where retries and corrective actions can change progress independently of timestamp order.

Task progress is a physical relation between states and a goal, so its prediction should be invariant to presentation order. This principle determines the design: the benchmark exposes order dependence, the data place the preferred physical state in both input positions, and supervision grounds the decision in visible task evidence. EgoTSR follows this chain from diagnosis to data construction and capability learning.

We focus on a specific claim: task progress in egocentric execution should be inferred from the relation between a visual state and the task goal, rather than from timestamp or input position. This claim yields three predictions. First, a model that has learned the comparison rule should preserve its physical preference when the same pair is presented in reverse order. Second, supervision that identifies task-relevant state changes and exposes both input positions should improve accuracy while reducing the forward--inverse Gap. Third, the same state-based rule should remain effective on failed, regressed, and recovered manipulation executions. SpatialLogic-Bench tests the first prediction, the CoT$\rightarrow$Tag framework tests the second, and the failure-aware experiments test the third.

\begin{figure*}[t]
    \centering
    \includegraphics[width=0.98\textwidth]
    {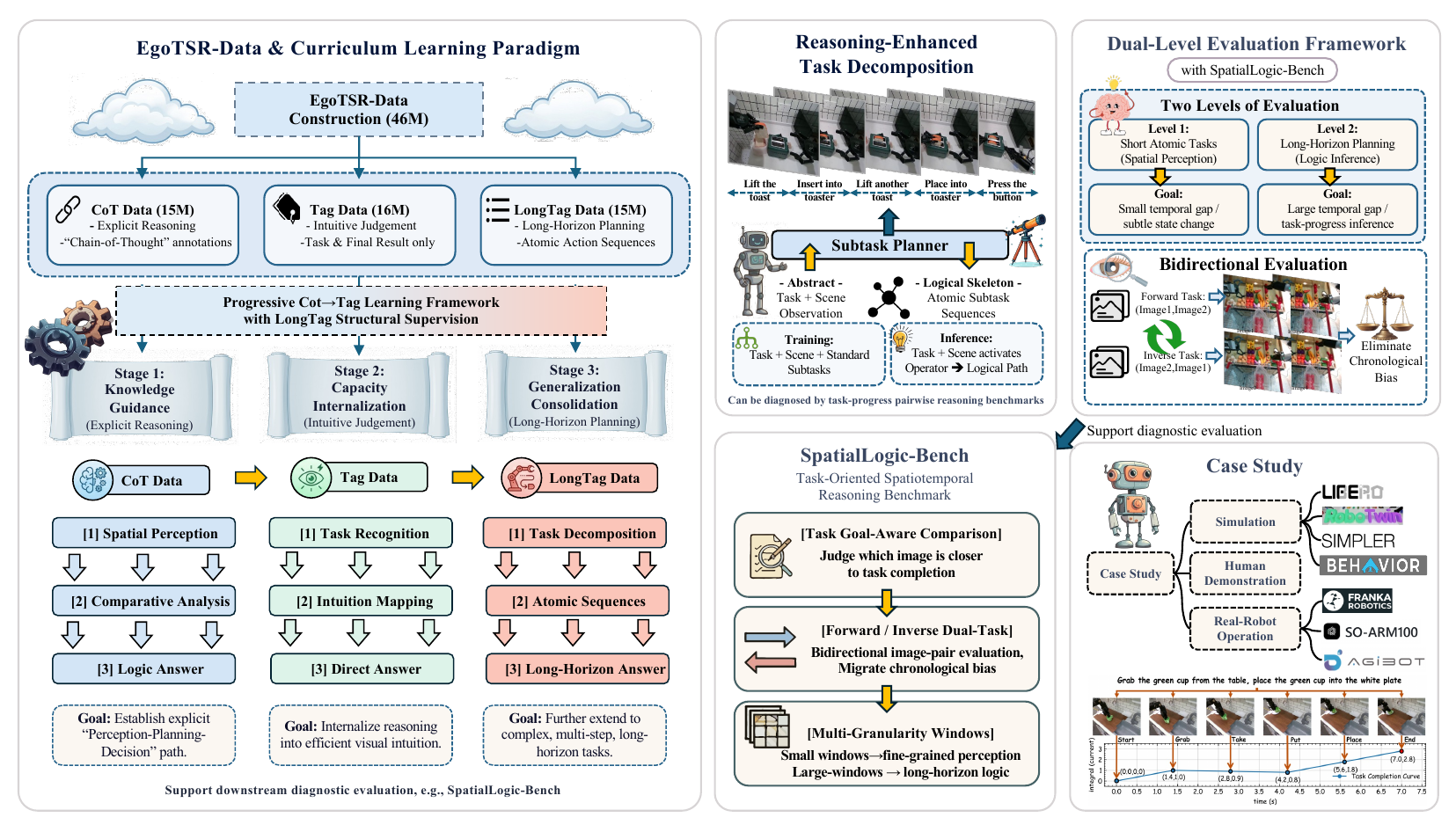}
    \caption{Overview of EgoTSR: SpatialLogic-Bench diagnoses input-order bias, EgoTSR-Data provides scalable state-comparison supervision, and CoT$\rightarrow$Tag learning produces an order-robust comparison rule.}
    \label{fig:overview}
\end{figure*}

SpatialLogic-Bench pairs every physical comparison with its order-swapped counterpart, turning input-order bias into a measurable forward--inverse Gap. Its short-horizon setting probes fine-grained state changes, and its long-horizon setting probes progress across omitted intermediate steps. The same diagnosis determines EgoTSR-Data: robot, human, and egocentric trajectories are converted into bidirectional supervision, and LongTag augments long-horizon pairs with an ordered subtask skeleton. The learning curriculum then uses CoT supervision to establish an evidence-grounded comparison path and Tag supervision to scale direct comparison training.

The results support this chain. Representative VLMs exhibit substantial forward--inverse gaps, while EgoTSR reaches $87.5\%$ short-horizon average accuracy (averaged over the eight reported short-horizon windows) and $92.4\%$ long-horizon average accuracy with a $0.1$-point long-horizon Gap. The core curriculum is trained on successful, approximately monotonic executions; failure-aware training extends the comparison rule to failure, regression, correction, and recovery trajectories, raising overall failure-aware accuracy from $64.1\%$ to $75.9\%$ and Recovery Accuracy from $58.6\%$ to $69.8\%$. External evaluations further show transfer to object-centric spatial reasoning and first-person interaction.

The main contributions are as follows:
\begin{itemize}
    \item \textbf{Goal-conditioned formulation and diagnosis.} We formulate egocentric task-progress understanding as comparison between two visual states conditioned on a task goal. Our proposed SpatialLogic-Bench tests this capability on short- and long-horizon manipulation trajectories and diagnoses whether predictions remain consistent after input-order reversal.
    \item \textbf{Scalable task-progress supervision.} EgoTSR-Data constructs bidirectional comparison pairs from manipulation and diverse first-person trajectories. LongTag retains ordered subtask structure for comparisons across omitted intermediate steps.
    \item \textbf{Progressive capability learning.} The CoT$\rightarrow$Tag curriculum combines evidence-grounded reasoning supervision with scalable label supervision, producing a comparison rule that remains stable across input orders.
\end{itemize}

\section{Related Work}

\subsection{Goal-Conditioned Progress Benchmarks}

Visual reasoning benchmarks span several levels of scene understanding. CLEVR tests compositional reasoning in controlled scenes~\cite{johnson2017clevr}, while GQA evaluates compositional questions about real images~\cite{hudson2019gqa}. ScanNet provides annotated indoor 3D reconstructions~\cite{dai2017scannet}, and Matterport3D supports RGB-D scene understanding~\cite{chang2017matterport3d}. For dynamic observations, Perception Test evaluates multimodal video perception~\cite{patraucean2023perceptiontest}, while TempCompass diagnoses temporal understanding in video language models~\cite{liu2024tempcompass}.

Related diagnostics examine language hallucination and visual illusion in HallusionBench~\cite{guan2024hallusionbench}, and state-change counterfactuals support procedure-aware video representations~\cite{kung2025statechange}. Task-progress understanding asks a complementary, goal-directed question: whether an observed state change brings an egocentric execution closer to its specified goal. We instantiate this question in manipulation, where progress can be grounded in a relation between two physical states and a task instruction.

Natural chronological presentation creates a strong prior: later frames are often closer to task completion. SpatialLogic-Bench makes the missing comparison explicit by evaluating the same physical pair in both orders. The resulting forward--inverse protocol links fine-grained spatial state perception, cross-step task-progress reasoning, and order robustness within one evaluation.

\subsection{Multimodal Chain-of-Thought Supervision for Reasoning}

Chain-of-Thought (CoT) prompting uses explicit intermediate steps to support arithmetic, commonsense, and symbolic reasoning~\cite{wei2022cot}. Zero-shot reasoning prompts can also elicit such steps without worked examples~\cite{kojima2022zeroshot}. In embodied visual understanding, room-object entity prompting organizes scene information for referring-expression reasoning~\cite{gao2024roomobject}. These approaches motivate structured reasoning, with the evidence and intermediate steps tailored to the task.

Multimodal CoT methods commonly use inference-time prompts to elicit intermediate analyses. Training-time reasoning supervision offers a complementary route: it can organize visual evidence into a reusable decision path. For egocentric task-progress comparison, that path should connect object-state changes to the task goal and remain stable after input-order reversal.

Curriculum learning provides a useful precedent for staging difficult reasoning supervision~\cite{bengio2009curriculum}. EgoTSR applies this principle to state comparison: CoT supervision links task-relevant evidence, state-change interpretation, and progress judgment, while Tag supervision scales the resulting comparison rule through label-only training.

\subsection{Task Progress Understanding in Egocentric Execution}

Task-progress understanding connects robot learning with procedural video understanding. VIP learns value-implicit representations from videos for reward computation~\cite{ma2022vip}. SayCan grounds language-based planning in robotic affordances~\cite{ahn2022saycan}, while Eureka generates reward code with language models~\cite{ma2023eureka}. Other work obtains rewards from vision-language models~\cite{chen2024vlmreward} or automates robot-learning tasks through generative simulation~\cite{xiong2024robogen}. These approaches connect visual or language representations to robot-learning objectives; our focus is the reliability of goal-conditioned state comparison.

Temporal order is a useful signal on successful demonstrations, but it does not define task progress on its own. Real manipulation executions include failures, regressions, interruptions, local disturbances, and ineffective actions, so the physical relation between state and goal must be distinguished from timestamp position. EgoTSR makes this distinction the central object of evaluation and supervision.

Long-horizon manipulation adds procedural structure to this comparison. The model must locate each state within an ordered chain of subtasks while interpreting local object relations. EgoTSR-Data organizes robot, human, and egocentric trajectories into a shared pairwise interface, and LongTag preserves the subtask skeleton for cross-step comparisons. Progressive CoT$\rightarrow$Tag learning then converts explicit evidence paths into scalable order-robust state comparison. Failure-aware evaluation measures how this rule transfers to recoveries and regressions.

\section{Problem Formulation}
\label{sec:formulation}

We operationalize egocentric task-progress understanding as goal-conditioned comparison between visual states. Let \(L_{\mathrm{task}}\) denote a natural-language task instruction and let \(I_a\) and \(I_b\) be two observations from the same task execution. The model must determine which observation is closer to successful completion:
\begin{equation}
f(L_{\mathrm{task}}, I_a, I_b) \rightarrow \{a,b\}.
\end{equation}
The output identifies the candidate state with greater task progress. Unlike action classification or video captioning, the objective is relational: the model must identify task-relevant spatial changes and map them to a goal-conditioned progress judgment. The same interface applies to robot-mounted cameras and human first-person video, while manipulation supplies the primary setting for controlled evaluation.

\subsection{Goal-Conditioned State Comparison}

For successful trajectories, we use the physical ordering of task states to assign supervision. Given a downsampled trajectory \(\tilde{V}=\{\tilde{I}_1,\ldots,\tilde{I}_M\}\) whose configuration progresses toward task completion, the ground-truth state for a pair \((\tilde{I}_{t_a},\tilde{I}_{t_b})\) is the state with the larger original time index:
\begin{equation}
\label{eq:zgt-rule}
z_{\mathrm{gt}}\bigl(\tilde{I}_{t_a},\tilde{I}_{t_b}\bigr) =
\begin{cases}
\tilde{I}_{t_b}, & t_a < t_b, \\
\tilde{I}_{t_a}, & t_a > t_b.
\end{cases}
\end{equation}
This rule applies to successful, approximately monotonic executions and defines the preferred physical state independently of its input position.

\subsection{Order-Robust Evaluation}

To separate state comparison from input-position heuristics, we evaluate the same physical pair in two orders. The forward and inverse queries are
\begin{equation}
q^{\mathrm{fwd}}=(L_{\mathrm{task}},\tilde{I}_t,\tilde{I}_{t+\Delta}),
\qquad
q^{\mathrm{inv}}=(L_{\mathrm{task}},\tilde{I}_{t+\Delta},\tilde{I}_t).
\end{equation}
The physical answer is unchanged while its input position is swapped. A model that compares task-relevant state changes preserves its physical judgment across the two orders. We measure this property with forward accuracy, inverse accuracy, their average, and the forward--inverse gap
\begin{equation}
\mathrm{Gap}=\left|\mathrm{Acc}_{\mathrm{fwd}}-\mathrm{Acc}_{\mathrm{inv}}\right|.
\end{equation}
The benchmark introduced next instantiates this formulation with multi-granularity state pairs. Pairwise comparison keeps the question within a task instance, accommodates different trajectory lengths and subtask decompositions, and allows trajectories recorded from different viewpoints to share one evaluation interface. Section~\ref{sec:ablation} also reports a single-image progress baseline for comparison.

\section{SpatialLogic-Bench}
\label{sec:spatiallogic}

\begin{figure*}[t]
    \centering
    \includegraphics[width=0.98\textwidth]
     {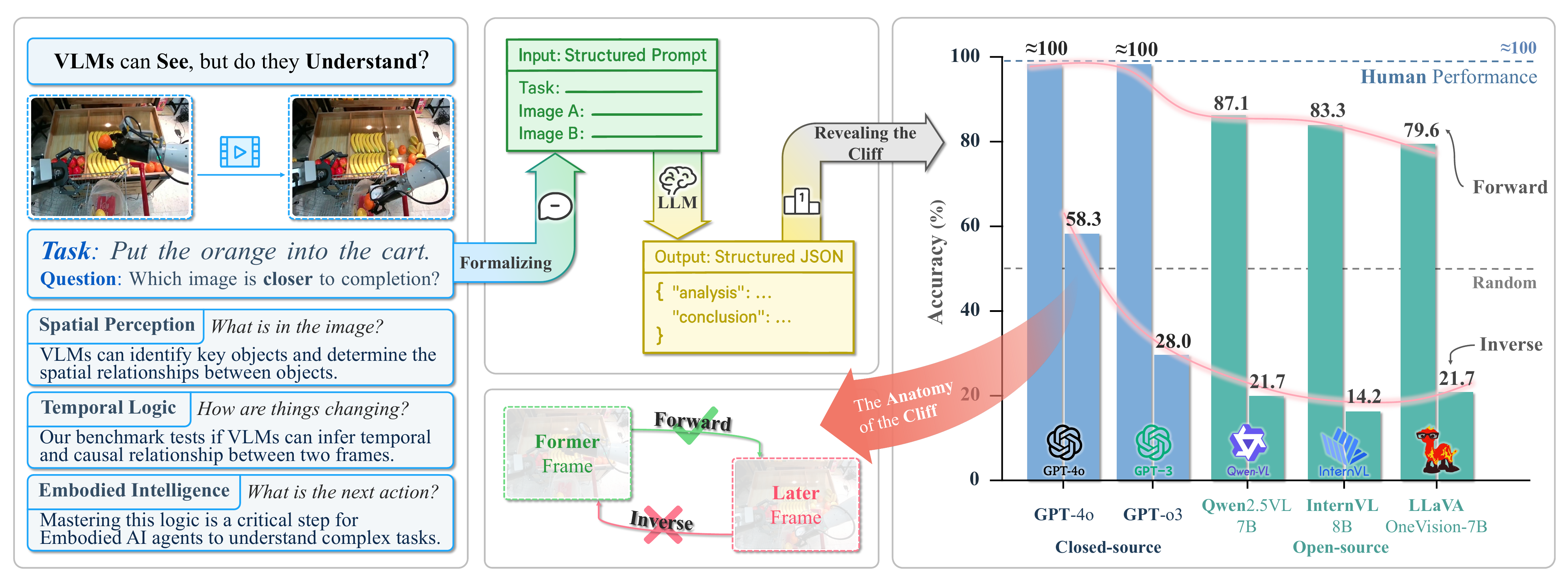}
     \caption{Illustration of SpatialLogic-Bench. The benchmark instantiates task-goal-aware state comparison and uses forward--inverse evaluation to diagnose input-order bias.}
     \label{fig:teaser}
\end{figure*}

\subsection{Benchmark Scope and Diagnostic Goal}

Most existing evaluations of Vision-Language Models (VLMs) focus on static spatial relations, action recognition, or general video understanding. SpatialLogic-Bench~\cite{yang2026spatiallogic} targets a task-specific relation underlying egocentric progress understanding: whether one observed state is closer to the task goal than another.

In manipulation tasks such as organizing, assembling, or transporting objects, the key evidence is often a change in configuration rather than a salient action category. Relevant differences include contact relations, relative position, object orientation, and local topology. SpatialLogic-Bench therefore evaluates goal-conditioned state comparison rather than generic video recognition.

Based on the formulation in Section~\ref{sec:formulation}, SpatialLogic-Bench instantiates the task with real manipulation trajectories and tests whether the comparison remains valid under input-order reversal. Manipulation serves as the primary diagnostic domain because it provides observable goal-directed changes and precise task descriptions. The benchmark measures two complementary capabilities: identifying local and task-relevant spatial state differences, and mapping those differences to goal-conditioned task-progress judgments.

Manipulation videos naturally correlate time with completion, making chronology an attractive shortcut. SpatialLogic-Bench exposes this shortcut by preserving the physical pair and reversing only its presentation order. Genuine progress reasoning should therefore remain stable across the two queries.

\subsection{Multi-Granularity Pair Construction}
\label{subsec:dual-task}

SpatialLogic-Bench is built from real manipulation trajectories, including large-scale real-world manipulation data such as AgiBot-World~\cite{agibot2025world}. Compared with synthetic data or static image collections, real videos contain continuous state evolution and task-relevant object interactions, making them well suited for evaluating goal-conditioned state comparison.

To obtain candidate frames with meaningful state differences, we first apply \(10\times\) temporal downsampling to the original videos, reducing camera jitter, short-term blur, and highly redundant neighboring frames. For each downsampled trajectory \(\tilde{V}=\{\tilde{I}_1,\tilde{I}_2,\ldots,\tilde{I}_M\}\), we construct pairs \((\tilde{I}_t,\tilde{I}_{t+\Delta})\) satisfying \(t+\Delta\leq M\). The temporal span \(\Delta\) controls the reasoning granularity: smaller spans emphasize local state changes, whereas larger spans cover more unobserved intermediate steps and emphasize cross-step task-progress reasoning. We evaluate spans \(\Delta\in\{5,6,7,8,9,10,11,\geq12\}\), forming a spectrum from local state perception to long-horizon task reasoning.

Multi-granularity sampling alone is insufficient for diagnosing whether a model truly understands task progress. As long as the inputs follow natural chronology, the model may still exploit the shallow heuristic that later states are more likely to be closer to completion. SpatialLogic-Bench therefore introduces a dual-task paradigm based on forward and inverse temporal orders.

Following the order-robust formulation in Section~\ref{sec:formulation}, each physical pair is evaluated in both forward and inverse input orders. The inverse task swaps only the input positions of the two visual states while keeping the task goal and underlying physical states unchanged. The state closer to task completion therefore remains the same, while its position label changes from \(b\) to \(a\), or vice versa. This construction prevents fixed input position or natural temporal order from serving as a shortcut, forcing the model to judge from the relationship between the visual states and the task goal.

\subsection{Evaluation Protocol and Metrics}

SpatialLogic-Bench evaluates models along two dimensions: temporal scale and input order. According to the window span, we divide samples into short-horizon and long-horizon subsets. Short-horizon samples mainly evaluate fine-grained local state discrimination, while long-horizon samples evaluate the model's ability to reason about task progress across omitted intermediate steps.

For each state pair, we report model accuracy under both forward and inverse input orders. The two scores measure task-progress judgment under the natural and order-swapped presentations, while their difference quantifies sensitivity to presentation order.

For each state pair, we report three metrics jointly: average accuracy \(\mathrm{Acc}_{\mathrm{avg}}=(\mathrm{Acc}_{\mathrm{fwd}}+\mathrm{Acc}_{\mathrm{inv}})/2\), the forward--inverse Gap defined in Section~\ref{sec:formulation}, and worst-case accuracy \(\mathrm{Acc}_{\mathrm{worst}}=\min(\mathrm{Acc}_{\mathrm{fwd}},\mathrm{Acc}_{\mathrm{inv}})\). Gap measures sensitivity to input-order perturbation, whereas \(\mathrm{Acc}_{\mathrm{worst}}\) measures performance under the harder ordering. For the failure-aware split, Failure Avg. is the overall accuracy across all valid forward--inverse evaluation items. This item-level aggregation complements the pair-type analysis reported below. In the current setting, SpatialLogic-Bench contains \(28{,}250\) evaluation samples covering diverse task categories and physical environments, making task-progress understanding and temporal-order dependence directly comparable.

\section{EgoTSR-Data Construction}
\label{sec:data}

\begin{figure*}[t]
    \centering
    \includegraphics[width=0.98\textwidth]{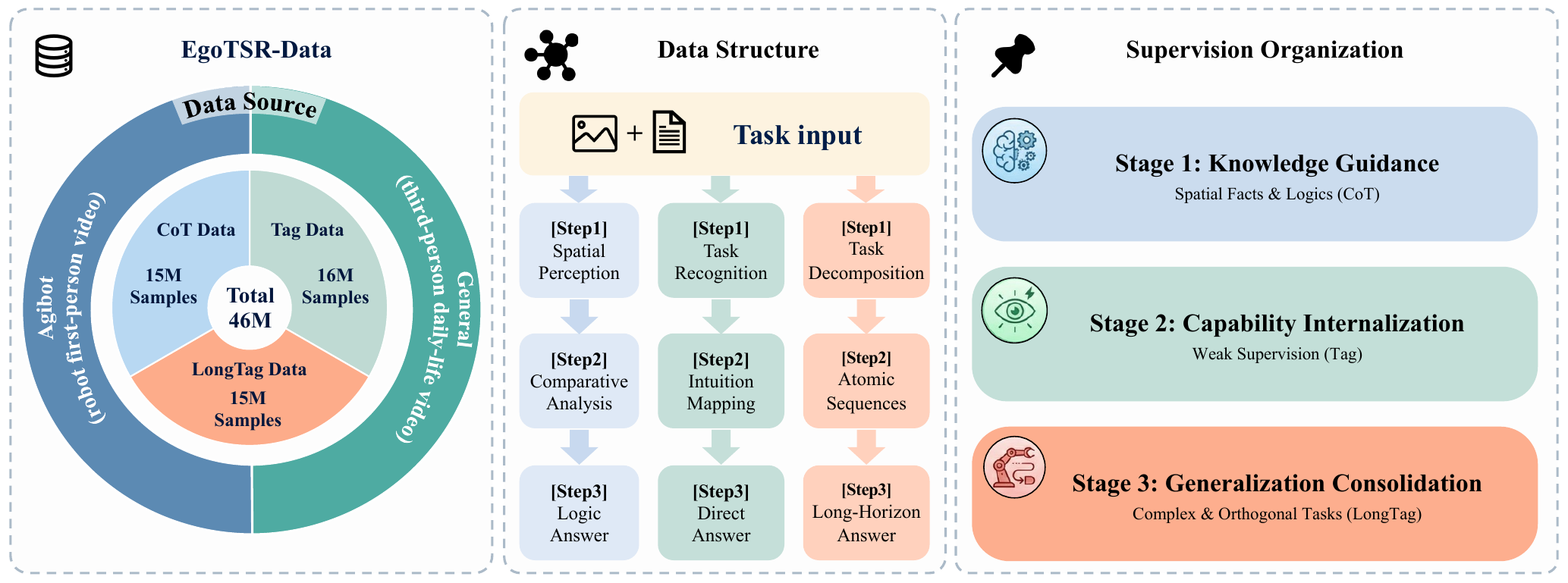}
    \caption{EgoTSR-Data construction and supervision organization. Successful robot, human, and egocentric episodes are converted into goal-conditioned state-comparison pairs and organized into CoT, Tag, and LongTag formats.}
    \label{fig:data}
\end{figure*}

\subsection{Data Sources and Pair-Level Objective}
\label{sec:data-source}

Existing video datasets typically provide action labels, captions, or temporally ordered clips. EgoTSR-Data converts these trajectories into explicit supervision for deciding which visual state is closer to a specified task goal. Its basic unit is a task-conditioned state-comparison pair.

We adopt the same label assignment as SpatialLogic-Bench (Eq.~\ref{eq:zgt-rule}): \(z_{\mathrm{gt}}\) is the frame with the larger original time index from a successful trajectory. The bidirectional construction in Section~\ref{sec:bidirectional} (Eqs.~\ref{eq:bidirectional-fwd}--\ref{eq:bidirectional-inv}) swaps the input slot occupied by \(z_{\mathrm{gt}}\) across paired samples.

Our data span robot manipulation trajectories from AgiBot World~\cite{agibot2025world} and Open X-Embodiment~\cite{oneill2024openx}, human egocentric operations from EPIC-KITCHENS~\cite{damen2018epickitchens55} and its 100-hour extension~\cite{damen2022epickitchens100}, and broader egocentric activities from Ego4D~\cite{grauman2022ego4d}. Robot trajectories supply controlled object transitions, human egocentric videos add viewpoint variation, and general ego datasets contribute long-tail action--object combinations. All are mapped to the same goal-conditioned comparison objective. The standard training pool contains successful, approximately monotonic episodes, for which temporal order provides a valid operational preference. Models trained on this pool are evaluated on both order-swapped pairs and non-monotonic executions. Before pair generation, each raw trajectory or video episode is assigned to a single split, and all derived pairs remain within that split. This trajectory-level organization prevents train--test leakage from neighboring frames sampled from the same continuous episode; for long-horizon samples, the associated subtask structures remain with the source trajectory.

For pair generation, we apply $10\times$ temporal downsampling to reduce redundant frames, camera jitter, and short-term blur. From a downsampled trajectory $\tilde{V}=\{\tilde{I}_1,\ldots,\tilde{I}_M\}$, we sample state pairs with multiple temporal spans:
\begin{equation}
\mathcal{P}(\tilde{V})=\{(\tilde{I}_t,\tilde{I}_{t+\Delta},L_{\mathrm{task}},z_{\mathrm{gt}})\mid t+\Delta\le M,\Delta\in\mathcal{D}_{\mathrm{train}}\},
\end{equation}
where $z_{\mathrm{gt}}$ denotes the physical state closer to completion, determined by the successful-trajectory rule above. Short-span pairs emphasize local spatial changes such as contact, relative position, placement, and orientation; long-span pairs require reasoning across omitted intermediate steps. The core training pool is formed only from successful trajectories assigned to the training split, with validation and test pairs generated independently from their own trajectory splits.

\subsection{Dataset Partition and Long-Horizon Structure}
\label{sec:data-partition}
EgoTSR-Data is organized into three subsets with different temporal scales and distributional roles. Short-Horizon Data contains pairs with small temporal spans, where decisive evidence is often a fine-grained spatial change rather than a salient action category. It supports local state discrimination, such as whether an object has contacted a target, whether its orientation better matches the task, or whether its relative position has moved toward the desired configuration. Long-Horizon Data contains pairs separated by larger temporal spans and often multiple hidden intermediate steps. Its challenge is to determine where each candidate state lies in the overall task procedure, not merely to identify a local visual difference. General Ego Data introduces first-person viewpoint variation and long-tailed action--object combinations that support transfer beyond scripted robot videos.

To provide structure for long-horizon comparison, a Subtask Planner predicts an ordered skeleton $S=\{s_1,s_2,\ldots,s_K\}$ from the initial observation $I_0$ and task instruction $L_{\mathrm{task}}$. Each $s_k$ is an atomic subtask and the ordering encodes procedural dependencies. The planner uses the Qwen2.5-VL-7B backbone and is fine-tuned on structured training triples $(I_0,L_{\mathrm{task}},S_{\mathrm{gt}})$. Ground-truth subtask metadata is confined to the training split: it supervises the planner and supplies LongTag context for training samples. For validation and test episodes, the planner produces $\hat{S}$ from $I_0$ and $L_{\mathrm{task}}$; neither ground-truth subtask metadata nor the pairwise comparison label is provided to the comparator. This split-specific inference rule is used for all reported LongTag results.

\subsection{CoT, Tag, and LongTag Supervision}
\label{subsec:supervision_formats}

EgoTSR-Data defines three supervision formats. CoT supervision attaches explicit reasoning content to each state-comparison pair. The reasoning identifies task-relevant evidence, interprets the state change, and explains the resulting progress preference. Short-horizon reasoning emphasizes local spatial evidence; long-horizon reasoning also relates each state to the task procedure. CoT annotations are produced with a teacher-assisted pipeline that first elicits spatial differences and then the final comparison label. Each pair is queried in both orders, and inconsistent physical judgments are filtered.

Tag supervision retains the final comparison label and scales direct state-comparison training. The model learns to invoke the evidence structure established by CoT without an explicit reasoning trace.

LongTag augments Tag supervision with the subtask skeleton $S$ for long-horizon samples. It preserves the task-logic structure needed to compare states across multiple subtasks. EgoTSR-Data therefore contains five data-supervision combinations: short CoT, short Tag, long CoT, LongTag, and general Tag, as summarized in Table~\ref{tab:data_organization}. All subsets share the comparison objective while varying in temporal span, supervision density, and structured task context.

\begin{table}[t]
\centering
\caption{Organization of EgoTSR-Data. $S$ denotes the subtask-level skeleton from metadata or the Subtask Planner.}
\label{tab:data_organization}
\footnotesize
\setlength{\tabcolsep}{4.5pt}
\renewcommand{\arraystretch}{1.0}
\begin{tabular}{@{}llll@{}}
\toprule
Subset & Horizon & Format & Primary role \\
\midrule
Short CoT & short & CoT & local evidence path \\
Short Tag & short & Tag & scalable local comparison \\
Long CoT & long & CoT & structured progress reasoning \\
LongTag & long & Tag + $S$ & long-horizon state comparison \\
General Tag & variable & Tag & long-tail generalization \\
\bottomrule
\end{tabular}
\end{table}

\subsection{Bidirectional Construction and Dataset Scale}
\label{sec:bidirectional}

EgoTSR-Data uses bidirectional construction to align the training interface with the order-robust evaluation. Suppose $I^+$ is physically closer to task completion than $I^-$ under $L_{\mathrm{task}}$. We construct both the natural-order sample
\begin{equation}
\label{eq:bidirectional-fwd}
x^{\mathrm{fwd}}=(I_a=I^-,I_b=I^+,L_{\mathrm{task}},y_{\mathrm{gt}}=b,C^{\mathrm{fwd}}),
\end{equation}
and the inverse-order sample
\begin{equation}
\label{eq:bidirectional-inv}
x^{\mathrm{inv}}=(I_a=I^+,I_b=I^-,L_{\mathrm{task}},y_{\mathrm{gt}}=a,C^{\mathrm{inv}}).
\end{equation}
The physical answer remains $I^+$ in both cases, while the input-position label changes. Tag samples use no additional context; LongTag samples retain the same subtask skeleton $S$; and CoT samples flip position references consistently with the image order. The resulting pair teaches the model to compare both states relative to the task goal.

The construction engine provides a large candidate pool from the core trajectories, with realized training subsets sampled for the CoT, Tag, and LongTag stages. We report split-level counts that distinguish unique physical pairs from their bidirectional presentations and source images. General ego data, including EPIC-KITCHENS-55 and EPIC-KITCHENS-100~\cite{damen2018epickitchens55,damen2022epickitchens100}, increases long-tail coverage through varied viewpoints and action--object state transitions.

\section{EgoTSR Learning Framework}
\label{sec:framework}

\begin{figure*}[t]
    \centering
    \includegraphics[width=0.98\textwidth]{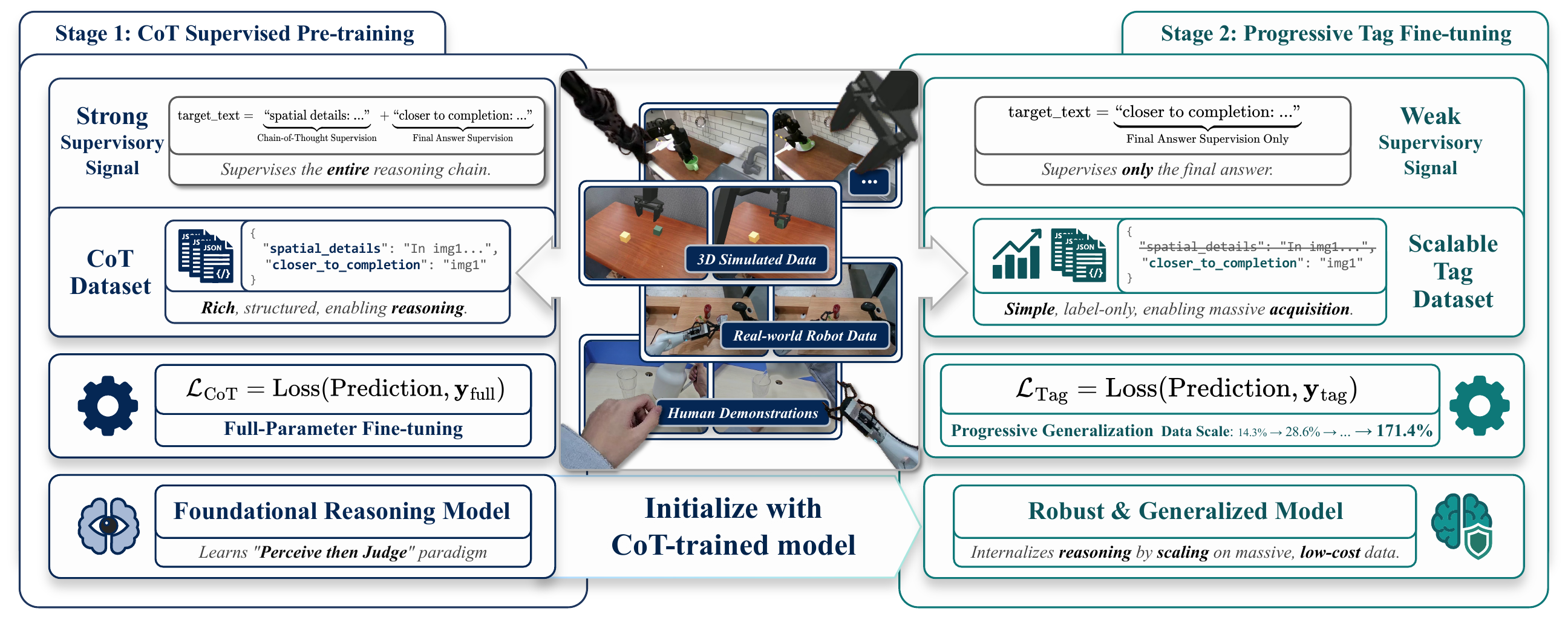}
    \caption{Progressive CoT$\rightarrow$Tag learning framework. CoT supervision establishes evidence-grounded reasoning paths, and scalable Tag supervision transfers them into order-robust state comparison.}
    \label{fig:framework}
\end{figure*}

\subsection{Overview}

The benchmark diagnosis determines the learning design. EgoTSR trains all data subsets with one goal-conditioned comparison objective through a progressive CoT$\rightarrow$Tag curriculum. CoT supervision establishes task-relevant evidence paths, and label-only supervision transfers those paths into direct comparison behavior across short, long, and general-ego data.

The curriculum separates explicit evidence supervision from scalable label training. CoT directs learning toward object-state changes, spatial relations, and their task meaning; Tag then scales the same decision rule across larger pair collections.

Given a task instruction \(L_{\mathrm{task}}\) and a candidate state pair \((I_a,I_b)\), the model organizes task-relevant spatial differences and their relation to the task goal, then predicts the input position \(\hat{y}\in\{a,b\}\) closer to task completion. Tag samples provide the comparison label, LongTag samples add the subtask skeleton \(S\), and CoT samples supervise the evidence path during training.

The CoT stage supervises the intermediate reasoning path; the Tag stage optimizes the final comparison label so that the same ability operates without an explicit reasoning trace.

\subsection{CoT Supervision for Reasoning Path Construction}

CoT supervision establishes an evidence path for goal-conditioned state comparison. For each pair, the model identifies task-relevant visual differences, interprets their relation to the manipulation goal, and produces the corresponding progress preference.

Following the CoT supervision format defined in Section~\ref{subsec:supervision_formats}, let the target reasoning text be \(C_{\mathrm{cot}}=(c_1^*,\ldots,c_T^*)\). Its tokens describe task-relevant evidence, the interpretation of the state change, and the resulting progress judgment. The corresponding token-level supervision is
\begin{equation}
\mathcal{L}_{\mathrm{reason}}
=-
\frac{1}{T}\sum_{t=1}^{T}
\log p_\theta\!\left(c_t^*\mid I_a,I_b,L_{\mathrm{task}},c_{<t}^*\right).
\end{equation}
The final comparison label is supervised together with the reasoning target:
\begin{equation}
\mathcal{L}_{\mathrm{CoT}}
=
 -\log p_\theta\!\left(y_{\mathrm{gt}}\mid I_a,I_b,L_{\mathrm{task}},C_{\mathrm{cot}}\right)
 +\lambda_{\mathrm{cot}}\mathcal{L}_{\mathrm{reason}},
\end{equation}
where \(y_{\mathrm{gt}}\) is the ground-truth comparison label and \(\lambda_{\mathrm{cot}}\) controls the contribution of intermediate reasoning supervision. This stage aligns the prediction with task-relevant evidence, state-change interpretation, and goal-conditioned progress judgment.

\subsection{Tag Supervision for Direct Comparison}

The Tag stage transfers the evidence path into direct state comparison. Given the state pair and task goal, label-only training supplies a scalable decision signal for learning a stable preference.

The Tag stage uses the Tag and LongTag formats defined in Section~\ref{subsec:supervision_formats}, covering short Tag, LongTag, and general Tag data. In LongTag, \(S\) supplies structured task context that preserves long-horizon task logic during label-only training.

The Tag-stage objective is
\begin{equation}
\mathcal{L}_{\mathrm{Tag}}
=
\mathcal{L}_{\mathrm{ans}}^{\mathrm{tag}},
\end{equation}
where \(\mathcal{L}_{\mathrm{ans}}^{\mathrm{tag}}\) denotes the prediction loss for the final comparison label. Applied over larger and more diverse pair collections, this objective consolidates the evidence-grounded comparison rule into direct predictions.

\subsection{Progressive Training Procedure}

EgoTSR is trained in the order of CoT$\rightarrow$Tag. The first stage uses the short- and long-horizon CoT subsets to construct explicit reasoning paths, and the second stage uses the short Tag, LongTag, and general ego Tag subsets to transfer the learned comparison rule to direct prediction. Denote the two supervision collections by \(\mathcal{D}_{\mathrm{cot}}\) and \(\mathcal{D}_{\mathrm{tag}}\), respectively.
The first-stage optimization is formulated as
\begin{equation}
\theta^{(1)}
=
\arg\min_{\theta}
\sum_{x_i\in\mathcal{D}_{\mathrm{cot}}}
\mathcal{L}_{\mathrm{CoT}}(x_i;\theta).
\end{equation}
This stage encourages the model to establish a judgment process grounded in task-relevant visual evidence.

The second stage is initialized from \(\theta^{(1)}\) and optimized as
\begin{equation}
\theta^{*}
=
\arg\min_{\theta}
\sum_{x_i\in\mathcal{D}_{\mathrm{tag}}}
\mathcal{L}_{\mathrm{Tag}}(x_i;\theta),
\quad
\theta \text{ initialized by } \theta^{(1)}.
\end{equation}

This stage converts the evidence-grounded comparison rule into direct prediction. Short, long, and general-ego data enter the same CoT$\rightarrow$Tag pathway with their respective temporal spans and structured contexts. The resulting model is evaluated on order-swapped pairs and on failure, regression, correction, and recovery trajectories.

\section{Experiments}
\label{sec:experiments}

The experiments evaluate the proposed interface for egocentric task-progress understanding from four complementary perspectives: input-order bias, comparison accuracy, the contribution of supervision and structural context, and transfer to non-monotonic executions, additional backbones, and first-person visual capabilities. The supplementary material provides per-window results, construction details, implementation settings, and additional qualitative examples.

\subsection{Experimental Setup}

\paragraph*{Benchmark.}
Our core experiments are conducted on SpatialLogic-Bench. The benchmark uses real manipulation trajectories as a controlled testbed for egocentric task-progress understanding and constructs task-conditioned state-comparison samples from them. Given a task instruction and two candidate visual states, the model must determine which state is closer to task completion. According to the temporal window span, the evaluation is divided into short-horizon and long-horizon subsets, which respectively assess fine-grained local state discrimination and cross-step task-progress reasoning. SpatialLogic-Bench also constructs both forward and inverse input orders for each physical pair, enabling explicit diagnosis of input-order bias.

\paragraph*{Metrics.}
We report \(\mathrm{Acc}_{\mathrm{avg}}\), Gap, and \(\mathrm{Acc}_{\mathrm{worst}}\) as defined in Section~\ref{sec:spatiallogic} on both short-horizon and long-horizon subsets. For non-monotonic trajectories, we additionally report Recovery Acc., which measures whether a recovered state is correctly judged as closer to completion than its preceding failed state, and Regression Error, which measures how often a regressed or failed state is incorrectly preferred over an earlier, better state. Higher Recovery Acc. is better, whereas lower Regression Error is better.

Evaluation counts distinguish source videos, unique physical pairs, and their forward/inverse ordered presentations. Candidate pairs are generated only after trajectory-level partitioning, so adjacent frames and reordered versions of the same physical relation cannot cross training and evaluation splits.

\begin{table*}[t]
\centering
\caption{Comprehensive performance on SpatialLogic-Bench. Short-horizon Avg. averages the eight listed window categories, except for Qwen3.5, for which only windows 5--10 were measured. Dashes denote unmeasured settings.}
\label{tab:main_spatiallogic}
\scriptsize
\setlength{\tabcolsep}{1.8pt}
\renewcommand{\arraystretch}{1.00}
\resizebox{\textwidth}{!}{
\begin{tabular*}{\textwidth}{@{\extracolsep{\fill}}lcccccccccccccccc@{}}
\toprule
\multirow{3}{*}{Model} 
& \multicolumn{9}{c}{Level 1: Short-Horizon Acc. (\%)} 
& \multicolumn{7}{c}{Level 2: Long-Horizon Acc. (\%)} \\
\cmidrule(lr){2-10} \cmidrule(lr){11-17}
& \multicolumn{8}{c}{Frame interval \((\times 10)\)} 
& \multirow{2}{*}{Avg.} 
& \multicolumn{3}{c}{Forward} 
& \multicolumn{3}{c}{Inverse} 
& \multirow{2}{*}{Avg.} \\
\cmidrule(lr){2-9} \cmidrule(lr){11-13} \cmidrule(lr){14-16}
& 5 & 6 & 7 & 8 & 9 & 10 & 11 & $\geq 12$
& 
& S & M & L
& S & M & L
& \\
\midrule
\multicolumn{17}{l}{\textit{Closed-source VLMs}} \\
GPT-4o-mini~\cite{openai2024gpt4omini}
& 70.0 & 70.3 & 72.8 & 67.7 & 60.0 & 67.5 & 67.7 & 70.3 & 68.3
& 64.9 & 69.0 & 85.7 & 34.6 & 30.3 & 42.5 & 55.0 \\
GPT-4o~\cite{openai2024gpt4o}
& 71.1 & 71.7 & 76.7 & 77.8 & 79.2 & 79.2 & 79.2 & 79.2 & 76.8
& 56.6 & 61.2 & 69.7 & 57.3 & 51.5 & 54.6 & 58.2 \\
GPT-o3~\cite{openai2025o3}
& 67.2 & 61.4 & 59.6 & 61.6 & 59.6 & 64.0 & 62.0 & 66.6 & 62.8
& 71.9 & 64.7 & 73.1 & 69.6 & 66.7 & 68.6 & 69.0 \\
Gemini-2.5-Pro~\cite{comanici2025gemini25}
& 71.3 & 75.0 & 72.8 & 75.0 & 77.2 & 80.9 & 89.7 & 100.0 & 80.2
& 74.1 & 78.1 & 64.0 & 63.2 & 42.1 & 57.5 & 62.0 \\
Doubao-1.5-Pro~\cite{volcengine2025aiapplab}
& 64.0 & 76.0 & 56.0 & 46.0 & 56.9 & 86.0 & 90.0 & 96.0 & 71.4
& 68.4 & 58.3 & 58.8 & 45.2 & 48.2 & 29.4 & 52.0 \\
Seed-1.6~\cite{byteplus2026seed16}
& 66.2 & 66.8 & 70.6 & 73.3 & 83.2 & 82.9 & 86.8 & 98.3 & 78.5
& 46.5 & 46.3 & 28.6 & 65.4 & 59.4 & 52.0 & 49.2 \\
\midrule
\multicolumn{17}{l}{\textit{Open-source 2D VLMs}} \\
NVILA-8B~\cite{liu2024nvila}
& 46.5 & 52.0 & 46.0 & 56.1 & 50.0 & 50.0 & 47.5 & 55.0 & 50.4
& 77.8 & 80.5 & 78.4 & 21.9 & 19.0 & 16.0 & 49.8 \\
PaliGemma~\cite{beyer2024paligemma}
& 43.8 & 45.1 & 44.2 & 45.7 & 42.3 & 43.6 & 42.5 & 46.1 & 44.2
& 76.0 & 81.9 & 87.6 & 16.8 & 14.1 & 16.0 & 49.5 \\
Qwen2.5-VL-3B~\cite{qwen2025vl}
& 51.3 & 54.2 & 57.1 & 58.8 & 54.5 & 59.6 & 59.6 & 66.7 & 57.7
& 24.0 & 17.3 & 12.1 & 82.9 & 84.4 & 83.1 & 49.9 \\
Qwen2.5-VL-7B~\cite{qwen2025vl}
& 53.1 & 55.4 & 55.0 & 57.3 & 55.2 & 56.1 & 56.0 & 56.5 & 55.6
& 93.1 & 89.1 & 85.0 & 8.3 & 11.9 & 13.6 & 49.8 \\
Qwen3.5 (direct)~\cite{qwen2026qwen35}
& 52.7 & 53.7 & 53.5 & 54.4 & 54.4 & 54.1 & -- & -- & 53.8
& -- & -- & -- & -- & -- & -- & -- \\
InternVL-8B~\cite{chen2024internvl}
& 47.7 & 50.6 & 49.4 & 47.9 & 47.1 & 48.7 & 48.9 & 51.0 & 48.9
& 98.8 & 99.5 & 99.7 & 1.5 & 2.2 & 2.2 & 50.6 \\
DeepSeek-VL2~\cite{wu2024deepseekvl2}
& 64.4 & 66.9 & 65.2 & 67.8 & 67.0 & 69.1 & 77.4 & 81.0 & 69.9
& 76.0 & 82.4 & 87.6 & 17.1 & 14.1 & 16.0 & 49.7 \\
LLaVA-OneVision~\cite{chen2024llavaonevision}
& 54.8 & 55.4 & 53.5 & 49.8 & 52.1 & 50.6 & 52.7 & 50.4 & 52.5
& 73.7 & 80.0 & 86.4 & 19.2 & 13.8 & 18.5 & 49.3 \\
\midrule
\multicolumn{17}{l}{\textit{Open-source 3D VLMs}} \\
3D-LLM~\cite{hong2023_3dllm}
& 27.0 & 28.0 & 30.5 & 30.3 & 33.9 & 32.5 & 36.8 & 37.7 & 32.1
& 86.8 & 90.5 & 84.3 & 7.5 & 9.6 & 7.6 & 47.7 \\
LL3DA~\cite{chen2023ll3da}
& 36.0 & 33.9 & 32.9 & 34.5 & 36.9 & 33.3 & 39.6 & 36.0 & 35.4
& 24.0 & 17.6 & 12.7 & 83.2 & 85.6 & 83.7 & 50.4 \\
Chat-Scene~\cite{huang2024chatscene}
& 30.8 & 32.5 & 40.7 & 49.0 & 41.6 & 45.3 & 43.3 & 51.0 & 41.8
& 24.3 & 18.1 & 12.4 & 82.9 & 85.3 & 83.4 & 50.3 \\
3D-LLaVA~\cite{deng2025_3dllava}
& 43.2 & 46.1 & 46.4 & 47.9 & 49.6 & 48.8 & 48.0 & 47.9 & 47.2
& 11.3 & 11.8 & 12.4 & 86.2 & 86.5 & 85.9 & 48.1 \\
Video-3D LLM~\cite{zheng2025video3dllm}
& 50.0 & 50.2 & 50.7 & 51.1 & 49.4 & 50.4 & 51.1 & 49.3 & 50.3
& 25.2 & 22.5 & 20.8 & 74.6 & 79.9 & 78.3 & 50.2 \\
\midrule
\multicolumn{17}{l}{\textit{EgoTSR variants}} \\
EgoTSR-CoT
& 68.3 & 70.2 & 72.2 & 73.7 & 74.5 & 75.6 & 76.7 & 77.8 & 73.6
& 91.9 & 90.1 & 93.0 & 1.8 & 2.2 & 2.6 & 46.9 \\
EgoTSR-CoT$\rightarrow$Tag
& 82.4 & 84.8 & 86.2 & 87.1 & 87.7 & 88.6 & 88.6 & 88.7 & 86.8
& 55.0 & 49.0 & 58.7 & 62.3 & 54.9 & 54.9 & 55.8 \\
EgoTSR-CoT$\rightarrow$Tag w/ LongTag
& \textbf{80.8} & \textbf{86.3} & \textbf{89.7} & \textbf{88.8} & \textbf{89.8} & \textbf{87.2} & \textbf{89.0} & \textbf{88.2} & \textbf{87.5}
& \textbf{88.2} & \textbf{92.9} & \textbf{96.2} & \textbf{92.0} & \textbf{92.7} & \textbf{92.3} & \textbf{92.4} \\
\bottomrule
\end{tabular*}
}
\end{table*}

\paragraph*{Baselines.}
We compare EgoTSR with closed-source VLMs, open-source 2D VLMs, 3D-aware VLMs, and matched EgoTSR variants. Table~\ref{tab:main_spatiallogic} reports the model comparisons, including direct-inference results for Qwen3.5 over windows 5--10. The supplementary material reports window-specific LoRA results across backbones.

Unless otherwise stated, EgoTSR denotes the full model trained on the core successful-trajectory pool. The first stage uses short and long CoT data to establish explicit reasoning paths; the second uses short Tag, LongTag, and general Tag data to transfer the learned rule through scalable label supervision. All variants share the same optimization setting and forward--inverse protocol. Main results use full-parameter fine-tuning of Qwen2.5-VL-7B, with LoRA~\cite{hu2022lora} portability reported in the supplementary material. Training uses AdamW~\cite{loshchilov2017adamw}, cosine learning-rate scheduling~\cite{loshchilov2016sgdr}, and mixed precision; the appendix provides complete hyperparameters, hardware cost, and training duration. The separate failure-aware variant additionally includes annotated failure, regression, correction, and recovery trajectories.

Table~\ref{tab:main_spatiallogic} shows that the core EgoTSR model delivers the strongest performance at both temporal levels: $87.5\%$ short-horizon average accuracy, obtained by averaging the eight short-horizon window categories, and $92.4\%$ long-horizon average accuracy. Its long-horizon forward and inverse scores remain consistently high across all three difficulty groups. In contrast, several representative VLMs display pronounced input-order bias. Under direct inference over windows 5--10, Qwen3.5 averages $91.79\%$ forward accuracy but only $15.81\%$ inverse accuracy, producing a $75.98$-point Gap despite $53.80\%$ overall accuracy. Qwen2.5-VL-7B and InternVL-8B show the same qualitative asymmetry, confirming that chronological presentation can dominate task-progress judgments.

The training variants trace how this advantage emerges. CoT supervision raises short-horizon accuracy to $73.6\%$ by establishing task-relevant evidence paths. Adding Tag supervision increases it to $86.8\%$ and balances the two presentation orders. LongTag then supplies the procedural structure needed for long-horizon comparison.

EgoTSR-CoT$\rightarrow$Tag with LongTag reaches the best overall result, combining $87.5\%$ short-horizon accuracy with $92.4\%$ long-horizon accuracy. The matched LongTag ablation attributes this gain to the preserved subtask skeleton, which supports comparisons across omitted intermediate steps.

\paragraph*{Why the forward--inverse protocol matters.}
The paired evaluation changes one factor at a time: the physical states, task instruction, and visual content remain fixed, while their input positions are exchanged. This makes the Gap a property of the model's comparison behavior rather than a consequence of different examples or task distributions. The baseline pattern in Table~\ref{tab:main_spatiallogic} gives the protocol a clear interpretation. Natural-order performance can be high when chronology agrees with progress, yet the inverse score reveals whether the same model can preserve the physical preference after that cue is removed. EgoTSR's high scores in both directions show that the gain is not obtained by favoring one query format.

\paragraph*{From local evidence to procedural progress.}
The short- and long-horizon results also show how the task scales. Short pairs expose changes such as contact, placement, orientation, and relative position. Long pairs require the model to connect those local changes to a sequence of subtasks whose intermediate states are not presented. The common comparison interface allows the two settings to be evaluated with the same metrics, while LongTag supplies the additional procedural context needed in the long-horizon case. Thus, the benchmark measures one capability at two resolutions: recognizing which visible configuration is better for the goal and locating that configuration within the larger manipulation procedure.

\paragraph*{Evidence for a learned comparison rule.}
The combination of order-swapped evaluation, the CoT$\rightarrow$Tag curriculum, and the LongTag ablation provides converging evidence for the proposed mechanism. The benchmark establishes the behavioral signature, CoT supervision organizes the evidence used to make the judgment, and Tag training scales the resulting rule to a larger pair collection. The structural ablation then shows that the subtask skeleton is useful precisely when the temporal span crosses intermediate procedural steps. Together, these results show how task-relevant state changes are converted into a reusable, order-robust progress preference.

\subsection{Ablation and Analysis}
\label{sec:ablation}

The ablations isolate four sources of EgoTSR's performance: supervision path, structured context, pair-order construction, and computation. Accuracy, worst-order accuracy, and Gap jointly capture comparison quality and order robustness.

\begin{figure}[t]
    \centering
    \includegraphics[width=0.98\columnwidth]{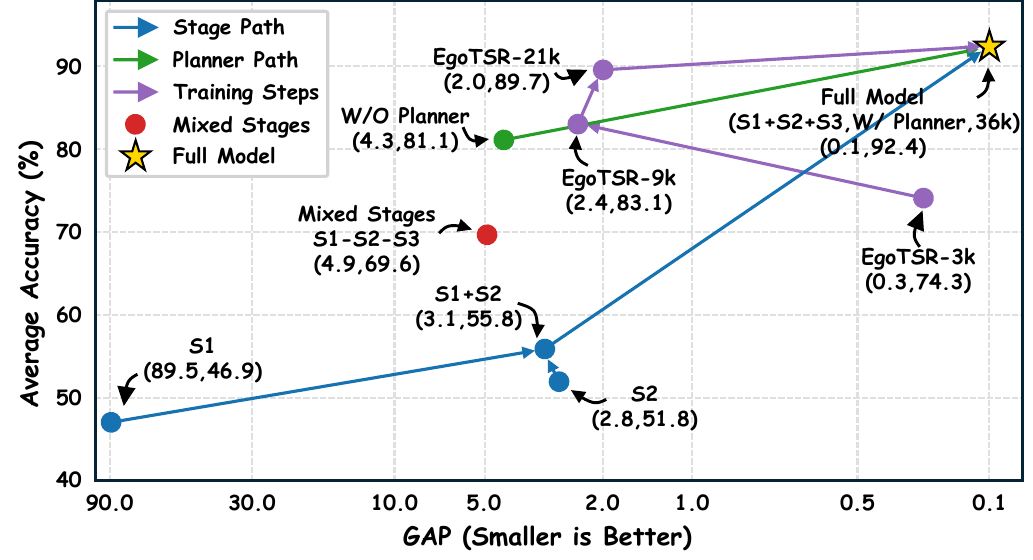}
    \caption{Ablation trajectories in the Accuracy--Gap space under the long-horizon protocol. Progressive training and LongTag move the model toward high accuracy and low input-order bias; the full EgoTSR model occupies the strongest operating point.}
    \label{fig:ablation_trajectory}
\end{figure}

Table~\ref{tab:ablation} shows that progressive CoT$\rightarrow$Tag training with LongTag achieves $92.4\%$ average accuracy, $92.3\%$ worst-order accuracy, and a $0.1$-point Gap. The sequential curriculum exceeds mixed CoT/Tag/LongTag training at the same data budget, establishing the value of the training order. The matched structural ablation adds a second result: supplying the subtask skeleton $S$ raises average accuracy from $81.1\%$ to $92.4\%$ and reduces the Gap from $4.3$ to $0.1$. CoT and Tag variants further identify the complementary roles of evidence grounding and order-balanced label training. Fig.~\ref{fig:ablation_trajectory} visualizes this progression in Accuracy--Gap space.

\begin{table}[t]
\centering
\caption{Ablation study of EgoTSR on long-horizon forward--inverse evaluation. \(\mathrm{Acc}_{\mathrm{worst}}=\min(\mathrm{Fwd},\mathrm{Inv})\) is reported jointly with Avg.\ and Gap to separate order-invariant reasoning from symmetric failure modes.}
\label{tab:ablation}
\small
\setlength{\tabcolsep}{2.1pt}
\renewcommand{\arraystretch}{0.98}
\begin{tabular}{@{}lccccc@{}}
\toprule
Method 
& Fwd $\uparrow$ 
& Inv $\uparrow$ 
& Avg.\ $\uparrow$ 
& Worst $\uparrow$
& Gap $\downarrow$ \\
\midrule
CoT only & 91.7 & 2.2 & 46.9 & 2.2 & 89.5 \\
Tag only & 50.4 & 53.2 & 51.8 & 50.4 & 2.8 \\
CoT$\rightarrow$Tag & 54.2 & 57.3 & 55.8 & 54.2 & 3.1 \\
Mixed CoT/Tag/LongTag & 67.2 & 72.1 & 69.6 & 67.2 & 4.9 \\
CoT$\rightarrow$Tag w/ LongTag
& \textbf{92.4} & \textbf{92.3} & \textbf{92.4} & \textbf{92.3} & \textbf{0.1} \\
\midrule
LongTag w/o \(S\) & 83.2 & 78.9 & 81.1 & 78.9 & 4.3 \\
LongTag w/ \(S\)
& \textbf{92.4} & \textbf{92.3} & \textbf{92.4} & \textbf{92.3} & \textbf{0.1} \\
\bottomrule
\end{tabular}
\end{table}

\paragraph*{Supervision path.}
On the matched short-horizon paired-order diagnostic subset, the full supervision path reaches $87.6\%$ forward accuracy and $85.1\%$ inverse accuracy. Its reported $86.4\%$ Avg. is their arithmetic mean, the worst-order accuracy is $85.1\%$, and the corresponding Gap is $2.5$ points (Table~\ref{tab:short_component}). CoT supplies task evidence, Tag balances the two presentation orders, and LongTag provides the final gain in both accuracy and stability. This diagnostic subset isolates supervision-path effects, while the main benchmark result aggregates the eight short-horizon window categories.

\begin{table}[t]
\centering
\caption{Supervision-path ablation on the matched short-horizon paired-order diagnostic subset. Worst is the lower of Fwd and Inv.}
\label{tab:short_component}
\small
\setlength{\tabcolsep}{0pt}
\renewcommand{\arraystretch}{0.98}
\begin{tabular*}{\columnwidth}{@{\extracolsep{\fill}}lrrrrr@{}}
\toprule
Variant & Fwd $\uparrow$ & Inv $\uparrow$ & Avg. $\uparrow$ & Worst $\uparrow$ & Gap $\downarrow$ \\
\midrule
Tag only & 69.4 & 66.2 & 67.8 & 66.2 & 3.2 \\
CoT only & 82.4 & 57.9 & 70.2 & 57.9 & 24.5 \\
CoT$\rightarrow$Tag & 84.1 & 78.2 & 81.2 & 78.2 & 5.9 \\
+ LongTag & \textbf{87.6} & \textbf{85.1} & \textbf{86.4} & \textbf{85.1} & \textbf{2.5} \\
\bottomrule
\end{tabular*}
\end{table}

\paragraph*{Pair-order construction.}
Pair-order construction affects comparison accuracy and order sensitivity differently (Table~\ref{tab:order_ablation}). Forward+inverse training reaches $81.2\%$ average accuracy, exceeding random-order training by $4.4$ points and improving worst-order accuracy from $74.6\%$ to $78.2\%$. Random-order training, however, has the smaller Gap at $4.3$ versus $5.9$ points. Bidirectional coverage therefore strengthens comparison accuracy in this diagnostic setting, whereas the lowest Gap emerges only when it is combined with progressive supervision and LongTag in EgoTSR-Full. The single-order variants show why coverage of both positions matters, with Gaps of $19.4$--$23.1$ points.

\begin{table}[t]
\centering
\caption{Pair-order ablation on the short-horizon paired-order diagnostic subset. Worst is the lower of Fwd and Inv.}
\label{tab:order_ablation}
\small
\setlength{\tabcolsep}{0pt}
\renewcommand{\arraystretch}{0.98}
\begin{tabular*}{\columnwidth}{@{\extracolsep{\fill}}lrrrrr@{}}
\toprule
Order strategy & Fwd $\uparrow$ & Inv $\uparrow$ & Avg. $\uparrow$ & Worst $\uparrow$ & Gap $\downarrow$ \\
\midrule
Forward only & 83.5 & 60.4 & 72.0 & 60.4 & 23.1 \\
Inverse only & 61.8 & 81.2 & 71.5 & 61.8 & 19.4 \\
Random order & 78.9 & 74.6 & 76.8 & 74.6 & 4.3 \\
Forward+inverse & 84.1 & 78.2 & 81.2 & 78.2 & 5.9 \\
EgoTSR-Full & \textbf{87.6} & \textbf{85.1} & \textbf{86.4} & \textbf{85.1} & \textbf{2.5} \\
\bottomrule
\end{tabular*}
\end{table}

\paragraph*{Simple baseline and efficiency.}
Table~\ref{tab:efficiency} positions EgoTSR across accuracy and training cost. The single-image baseline is trained on 100,000 single-frame items from the success-only training split. For each sampled successful pair, the earlier and later frames receive scalar progress targets of 0 and 1, respectively. At evaluation, each candidate frame is scored separately and the state with the larger predicted progress is selected; Avg. and Gap are then computed on the same paired-order diagnostic subset used by the pairwise models. This baseline reaches $70.6\%$ average accuracy at $0.18\times$ relative cost. LoRA raises accuracy to $82.1\%$ at $0.35\times$ cost, while full fine-tuning delivers $86.4\%$ average accuracy with a $2.5$-point Gap.

\begin{table}[t]
\centering
\caption{Accuracy, order sensitivity, and relative training cost on the matched short-horizon diagnostic subset.}
\label{tab:efficiency}
\normalsize
\setlength{\tabcolsep}{5.0pt}
\renewcommand{\arraystretch}{0.98}
\begin{tabular}{@{}lrrr@{}}
\toprule
Method & Avg. $\uparrow$ & Gap $\downarrow$ & Rel. cost $\downarrow$ \\
\midrule
Single-image regression & 70.6 & 9.4 & 0.18$\times$ \\
Tag only & 67.8 & 3.2 & 0.22$\times$ \\
LoRA & 82.1 & 3.9 & 0.35$\times$ \\
Full fine-tuning & \textbf{86.4} & \textbf{2.5} & 1.00$\times$ \\
\bottomrule
\end{tabular}
\end{table}

\begin{figure}[t]
    \centering
    \includegraphics[width=0.98\columnwidth]{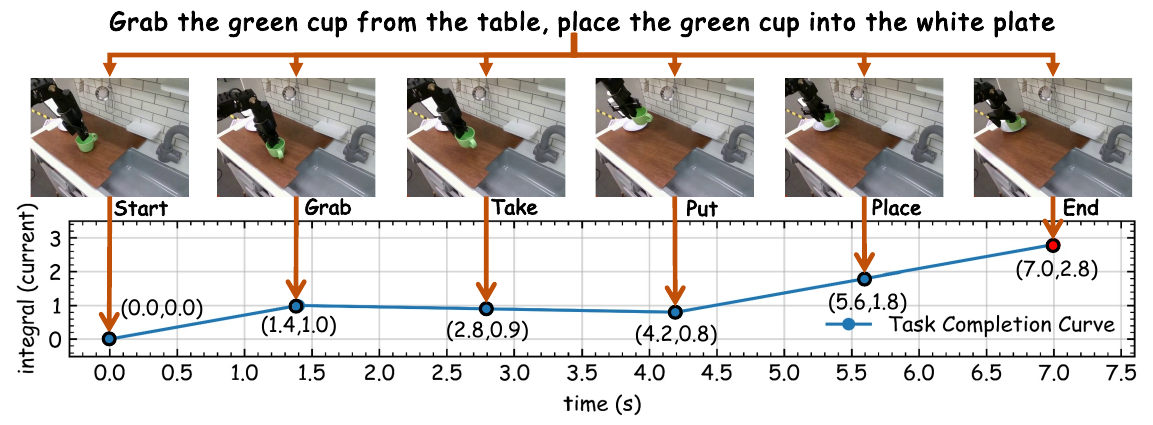}
    \caption{Qualitative task-completion curve on a long-horizon manipulation trajectory. The score changes most strongly around task-relevant state transitions.}
    \label{fig:case_curve}
\end{figure}

\paragraph*{Process-level qualitative analysis.}
We also aggregate pairwise predictions over unsegmented long-horizon videos to form a task-completion curve. In the representative \textit{grab-and-place} trajectory of Fig.~\ref{fig:case_curve}, the curve responds most strongly around grasping, transporting, and placing events and evolves smoothly during intermediate motion. The visualization connects pairwise decisions to process-level state evolution.

\subsection{Robustness to Non-Monotonic Trajectories}
\label{sec:failure_robustness}

The core EgoTSR-Success curriculum uses successful, approximately monotonic executions. To measure transfer to realistic execution variation, we construct a trajectory-disjoint split from BotFails~\cite{rolland2026failure} containing failed attempts, regressions, corrections, and recoveries. The primary failure-aware evaluation uses a fixed sliding-window span of 10 and contains 1,345 distinct state pairs, expanded to 2,690 forward--inverse evaluation items; recovery and regression pairs cover 9 and 24 source videos, respectively. Pair construction follows the video-level split, preserving trajectory independence across training and evaluation. Detailed pair counts and sliding-window sensitivity are reported in the supplementary material.

Table~\ref{tab:failure_robustness} compares EgoTSR-Success with EgoTSR-Failure-Aware on this fixed-Window-10 split. Adding annotated failure trajectories raises the overall failure-aware accuracy from $64.1\%$ to $75.9\%$, increases Recovery Acc. from $58.6\%$ to $69.8\%$, lowers Regression Error from $32.8\%$ to $23.4\%$, and reduces Gap from $13.5$ to $6.7$ points. The overall accuracy aggregates all valid ordered items in the split. EgoTSR-Failure-Aware also scores $84.1\%$ on the success test, extending the same comparator to retries and corrective actions.

\begin{table}[t]
\centering
\caption{Robustness on non-monotonic failure, regression, correction, and recovery trajectories. Succ. denotes success-only training and FA denotes failure-aware training. Failure Avg. is the overall accuracy across all valid ordered items in the fixed-Window-10 split. Failure Avg. and Recovery Acc. are higher-is-better; Regression Error and Gap are lower-is-better. Success Test reports accuracy on the original monotonic success split.}
\label{tab:failure_robustness}
\footnotesize
\setlength{\tabcolsep}{3.2pt}
\renewcommand{\arraystretch}{1.00}
\begin{tabular}{@{}lccccc@{}}
\toprule
Setting
& Fail. Avg. $\uparrow$
& Rec. $\uparrow$
& Reg. Err. $\downarrow$
& Gap $\downarrow$
& Succ. Test $\uparrow$ \\
\midrule
Qwen zero-shot & 54.8 & 46.5 & 44.7 & 18.9 & -- \\
EgoTSR-Succ. & 64.1 & 58.6 & 32.8 & 13.5 & \textbf{88.2} \\
EgoTSR-FA & \textbf{75.9} & \textbf{69.8} & \textbf{23.4} & \textbf{6.7} & 84.1 \\
\bottomrule
\end{tabular}
\end{table}

\paragraph*{Failure-type breakdown.}
Table~\ref{tab:failure_breakdown_main} breaks down the non-monotonic relations. The four categories represent normal--normal, failure--normal, normal--failure, and failure--failure pairs, respectively. EgoTSR reaches $76.8\%$ on Normal-progress, $76.6\%$ on Regression, and $73.6\%$ on Failure-internal, with Recovery Acc. at $69.8\%$. These category-level results characterize distinct transition types, while Table~\ref{tab:failure_robustness} reports the overall item-level accuracy.

\begin{table}[!htb]
\centering
\caption{Pair-type breakdown on the failure-aware split.}
\label{tab:failure_breakdown_main}
\normalsize
\setlength{\tabcolsep}{6pt}
\renewcommand{\arraystretch}{1.08}
\begin{tabular}{@{}lr@{}}
\toprule
Pair type & Accuracy (\%) $\uparrow$ \\
\midrule
Normal-progress & 76.8 \\
Recovery & 69.8 \\
Regression & 76.6 \\
Failure-internal & 73.6 \\
\bottomrule
\end{tabular}
\end{table}

\paragraph*{Sampling sensitivity.}
Table~\ref{tab:window_main} characterizes the accuracy and coverage induced by the sliding-window span. We use fixed Window 10 as the primary failure-aware construction, yielding 1,345 base pairs and the reported $75.9\%$ overall accuracy. Window 5, Window 20, and the adaptive 5+R10 policy are sensitivity settings; the latter focuses additional span on recovery and yields 1,292 pairs with $75.5\%$ overall accuracy.

\begin{table}[!htb]
\centering
\caption{Sliding-window sensitivity. The 5+R10 policy expands only recovery pairs to Window 10.}
\label{tab:window_main}
\normalsize
\setlength{\tabcolsep}{6pt}
\renewcommand{\arraystretch}{1.04}
\begin{tabular}{@{}lrrrr@{}}
\toprule
Window & Pairs & Avg. $\uparrow$ & Rec. $\uparrow$ & Gap $\downarrow$ \\
\midrule
5 & 1,198 & 74.8 & 67.6 & 7.2 \\
10 & 1,345 & 75.9 & 69.8 & 6.7 \\
20 & 1,824 & \textbf{78.6} & \textbf{71.1} & \textbf{5.9} \\
5+R10 & 1,292 & 75.5 & 69.4 & 6.8 \\
\bottomrule
\end{tabular}
\end{table}

\begin{figure*}[t]
    \centering
    \includegraphics[width=0.98\textwidth]{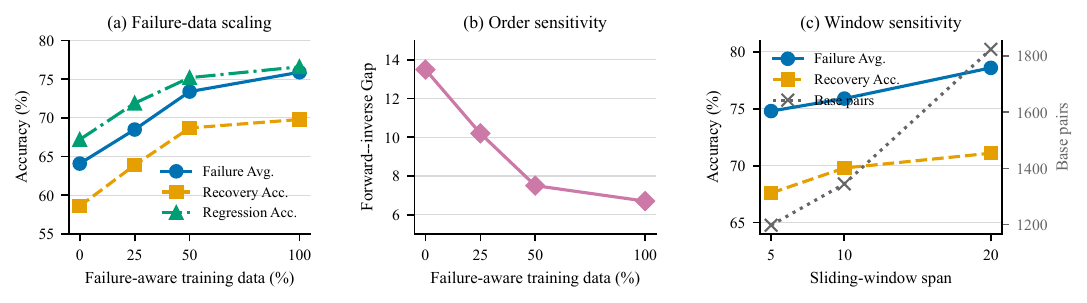}
    \caption{Quantitative analysis of failure-aware learning. (a) Failure, recovery, and regression accuracy as the amount of failure-aware training data increases. (b) The corresponding reduction in forward--inverse Gap. (c) Accuracy and candidate-pair growth under different sliding-window spans. Regression accuracy is one minus Regression Error.}
    \label{fig:failure_analysis}
\end{figure*}

\paragraph*{Failure-data scaling.}
Table~\ref{tab:failure_scale_main} and Fig.~\ref{fig:failure_analysis} show a smooth gain as the amount of failure-aware data increases. Failure Avg. rises from $64.1\%$ to $75.9\%$, Recovery Acc. rises from $58.6\%$ to $69.8\%$, and Gap falls from $13.5$ to $6.7$. The largest gains appear before the 50\% point, providing a practical operating range for failure-aware training.

\begin{table}[!htb]
\centering
\caption{Effect of failure-aware training-data scale.}
\label{tab:failure_scale_main}
\normalsize
\setlength{\tabcolsep}{4.2pt}
\renewcommand{\arraystretch}{1.02}
\begin{tabular}{@{}lrrrr@{}}
\toprule
Data & Fail. Avg. $\uparrow$ & Rec. $\uparrow$ & Reg. Err. $\downarrow$ & Gap $\downarrow$ \\
\midrule
0\% & 64.1 & 58.6 & 32.8 & 13.5 \\
25\% & 68.5 & 63.9 & 28.1 & 10.2 \\
50\% & 73.4 & 68.7 & 24.8 & 7.5 \\
100\% & \textbf{75.9} & \textbf{69.8} & \textbf{23.4} & \textbf{6.7} \\
\bottomrule
\end{tabular}
\end{table}

\paragraph*{Backbone portability.}
Table~\ref{tab:failure_backbone_main} applies the same failure-aware recipe across model families. Every backbone improves its failure-aware average and reduces its Gap, with gains from 4.4 to 11.8 points. The consistent trend shows that order-robust task-progress supervision transfers beyond the Qwen-7B implementation.

\begin{table}[t]
\centering
\caption{Backbone portability on failure-aware evaluation.}
\label{tab:failure_backbone_main}
\normalsize
\setlength{\tabcolsep}{4.2pt}
\renewcommand{\arraystretch}{1.02}
\begin{tabular}{@{}lrrrr@{}}
\toprule
Backbone & Base & EgoTSR & Gain $\uparrow$ & $\Delta$Gap $\downarrow$ \\
\midrule
Small VLM & 60.8 & 65.2 & +4.4 & -4.1 \\
Qwen-7B & 64.1 & \textbf{75.9} & \textbf{+11.8} & \textbf{-6.8} \\
InternVL & 66.7 & 74.2 & +7.5 & -5.6 \\
LLaVA-style & 62.5 & 68.9 & +6.4 & -4.8 \\
\bottomrule
\end{tabular}
\end{table}

\subsection{What the Diagnostic Reveals}
\label{sec:result_interpretation}

The forward--inverse protocol makes the central behavior directly measurable. A model can achieve a strong score in the observed order while still relying on the position of the two inputs; the swapped pair exposes whether its decision follows the physical state relation. The resulting Gap is therefore a direct measure of order robustness. This distinction matters most in long-horizon samples, where chronological distance is a convenient cue during successful executions but is not itself the task objective.

The three reported metrics expose complementary aspects of the same capability. Average accuracy measures comparison quality, Gap measures dependence on presentation order, and worst-order accuracy measures the reliability of the harder query direction. EgoTSR reaches 92.4\% average accuracy, 92.3\% worst-order accuracy, and a 0.1-point Gap in the long-horizon setting, placing all three quantities in the desired regime. Random-order training reduces order sensitivity, while the complete CoT$\rightarrow$Tag configuration also raises the underlying comparison accuracy. The result supports the interpretation that order invariance is learned as part of a task-grounded comparison rule rather than as a standalone output symmetry.

The ablations clarify how the learning path produces this regime. CoT supervision supplies an explicit evidence path for identifying task-relevant state changes; Tag supervision then turns that path into a scalable direct comparison rule. Their sequential combination improves both directions simultaneously. LongTag adds subtask structure for long-horizon trajectories, allowing the model to preserve intermediate task logic when the two observations are far apart. These components explain why the complete model improves accuracy and order stability together.

The failure-aware results extend the same comparison rule to realistic execution variation. Failure-aware training improves Failure Avg. by 11.8 points and Recovery Acc. by 11.2 points, while reducing Regression Error by 9.4 points and Gap by 6.8 points. The scaling curve rises smoothly as failure-aware data increase, while the fixed-Window-10 evaluation provides the primary measure of this transfer. The resulting model retains 84.1\% accuracy on the success test, showing that the learned rule transfers from canonical executions to retries, regressions, and repairs.

\subsection{Evaluation Controls}
\label{sec:evaluation_controls}

The evaluation is designed so that the measured change is attributable to the comparison rule. A physical pair is created once and its two ordered presentations are generated from the same images, task instruction, and ground-truth preference. The inverse query changes the input slots but does not introduce a new trajectory or a new visual example. This control gives the Gap a direct interpretation: the two accuracies differ only in how the same state relation is presented.

The data pipeline applies the same temporal preprocessing and pair-generation interface across sources. Videos are downsampled by $10\times$, candidate pairs are sampled at multiple spans, and all pairs from an episode remain in the episode's assigned split. The failure-aware split is also video-disjoint and is constructed separately from the successful training pool. These controls align the benchmark, training data, and robustness evaluation at the trajectory level, while keeping source-specific viewpoint and action distributions available for transfer analysis.

The main model uses Qwen2.5-VL-7B with full-parameter fine-tuning under the shared CoT$\rightarrow$Tag objective. The reported comparisons use the same forward--inverse protocol and metric definitions across variants. AdamW optimization, cosine scheduling, and mixed precision provide a common training setup; LoRA results quantify the lower-cost version in the efficiency analysis. Together, these choices make the principal comparisons differ in supervision path, structured context, pair-order coverage, or failure-aware data, rather than in the evaluation interface.

\subsection{External Generalization}

Table~\ref{tab:external_summary} shows that task-progress training transfers to complementary visual and spatial capabilities. EgoTSR-Tag obtains the strongest reported scores on both CV-Bench~\cite{tong2024cambrian} and MMSI-Bench~\cite{yang2026mmsi}, while EgoTSR leads the compared methods on VSI-Bench~\cite{yang2025thinkingspace} and the first-person Ego4D+EPIC-Kitchens interaction group~\cite{grauman2022ego4d,damen2022epickitchens100}. M1/M2 show selected submetrics, and Avg. reports the overall score for each evaluation.

The four groups probe distinct transfer paths. CV-Bench measures visual and depth cues, MMSI-Bench and VSI-Bench emphasize object-centric spatial reasoning, and the first-person group tests interaction understanding in unseen videos. The strongest gains occur on object-region relations and task-state comparisons, which are structurally close to the learned supervision. Table~\ref{tab:retention} further shows a stable aggregate capability profile after training: the overall score is 71.0, and temporal ordering improves by 1.1 points. Together, these results place EgoTSR's main contribution in order-robust task-progress reasoning while demonstrating useful transfer to related visual skills.

\begin{table}[t]
\centering
\caption{Compact external generalization results. Within the CV, MMSI, VSI, and Ego groups, M1/M2 denote Omni3D/Depth, Obj.-Obj./Obj.-Reg., Size/Rel.Dist., and E4D-L/EK-L, respectively. M1/M2 are selected submetrics; Avg. is the overall evaluation score.}
\label{tab:external_summary}
\normalsize
\setlength{\tabcolsep}{6pt}
\renewcommand{\arraystretch}{1.05}
\begin{tabular}{@{}llrrr@{}}
\toprule
Group & Model & M1 & M2 & Avg. \\
\midrule
\multirow{3}{*}{CV} & Qwen2.5-VL-7B~\cite{qwen2025vl} & 75.17 & 70.83 & 75.07 \\
 & EgoTSR-CoT & 73.00 & 73.33 & 72.34 \\
 & EgoTSR-Tag & \textbf{76.42} & \textbf{79.17} & \textbf{75.27} \\
\midrule
\multirow{3}{*}{MMSI} & Qwen2.5-VL-7B~\cite{qwen2025vl} & 24.5 & 29.4 & 25.9 \\
 & EgoTSR-CoT & 30.2 & 34.1 & 28.3 \\
 & EgoTSR-Tag & \textbf{33.0} & \textbf{36.5} & \textbf{29.6} \\
\midrule
\multirow{3}{*}{VSI} & LongVILA-8B~\cite{chen2024longvila} & 16.7 & 29.6 & 23.5 \\
 & InternVL2-2B~\cite{opengvlab2024internvl2} & 20.0 & 32.1 & 25.4 \\
 & EgoTSR & 21.9 & \textbf{47.1} & \textbf{29.3} \\
\midrule
\multirow{5}{*}{Ego} & Qwen2.5-VL-7B~\cite{qwen2025vl} & 50.2 & 46.1 & 49.2 \\
 & NVILA-8B~\cite{liu2024nvila} & 48.7 & 48.5 & 49.2 \\
 & LLaVA-OneVision~\cite{chen2024llavaonevision} & 50.6 & 49.7 & 50.5 \\
 & 3D-LLM~\cite{hong2023_3dllm} & 49.9 & 49.9 & 50.2 \\
 & EgoTSR & \textbf{59.0} & \textbf{57.0} & \textbf{57.5} \\
\bottomrule
\end{tabular}
\end{table}

\begin{table}[!htb]
\centering
\caption{Capability retention before and after goal-conditioned task-progress training.}
\label{tab:retention}
\normalsize
\setlength{\tabcolsep}{6pt}
\renewcommand{\arraystretch}{1.08}
\begin{tabular}{@{}lrrr@{}}
\toprule
Capability group & Before & After & Change \\
\midrule
General VQA & 73.1 & 71.8 & -1.3 \\
Spatial reasoning & 68.6 & 67.4 & -1.2 \\
Temporal ordering & 70.5 & \textbf{71.6} & +1.1 \\
Object-state QA & 74.0 & 73.2 & -0.8 \\
\midrule
Average & 71.6 & 71.0 & -0.6 \\
\bottomrule
\end{tabular}
\end{table}

Table~\ref{tab:retention} evaluates the capability profile before and after task-progress training. The aggregate score remains close at 71.6 versus 71.0, while temporal ordering improves by 1.1 points. This profile is consistent with targeted adaptation: the training objective strengthens temporal and state-comparison behavior while maintaining the broader visual capabilities measured by the held-out groups.

\subsection{Reading the State-Relation Results}
\label{sec:state_relation_reading}

Recovery remains the most demanding transition type because the model must connect a repaired configuration back to the task goal, whereas regression can often be identified from an immediate deterioration. The sampling, data-scale, and backbone studies nevertheless show a consistent pattern: broader failure-aware coverage improves overall and recovery accuracy while reducing both Regression Error and input-order Gap. Gains across all four backbone families further indicate that this behavior comes from the supervision signal rather than a model-specific effect.

\FloatBarrier
\subsection{Discussion and Future Work}
\label{sec:practical_discussion}

The pairwise interface serves as a state-assessment primitive for robotic systems. It can rank candidate checkpoints relative to a task goal, aggregate preferences into a task-completion curve, and flag inconsistent judgments when the same pair is queried in both orders. The failure-aware configuration extends this interface to retries, regressions, and recoveries, while the forward--inverse protocol provides an operational check on order stability. Integrating these preferences with temporal history, confidence estimation, or a controller is a natural next step toward checkpoint selection and execution monitoring.

This study keeps the measured capability precise: given a task instruction and two visual observations, the comparator estimates which state is closer to the goal. The benchmark emphasizes robotic manipulation, while the training data and external evaluation additionally cover human first-person activities. Future work can connect pairwise scores with visual-token attribution and representation consistency, add temporal memory, expand recovery coverage, and validate the comparator in a closed-loop manipulation protocol.

\section{Conclusion}

This paper studies egocentric task-progress understanding through goal-conditioned comparison between visual states. The formulation isolates a capability that chronological evaluation obscures: deciding which state is closer to the goal while remaining invariant to presentation order. SpatialLogic-Bench measures this capability on manipulation trajectories with matched forward and order-swapped queries, EgoTSR-Data supplies scalable supervision from robot and human egocentric executions, and the progressive CoT$\rightarrow$Tag curriculum turns evidence-grounded reasoning into a stable inference rule.

EgoTSR reaches 87.5\% short-horizon average accuracy and 92.4\% long-horizon average accuracy, with a 0.1-point long-horizon Gap. Failure-aware supervision raises Failure Avg. by 11.8 points and Recovery Acc. by 11.2 points, while external evaluations show transfer to visual, spatial, and first-person interaction capabilities. These results establish order-robust goal-conditioned state comparison as an explicit and transferable formulation for egocentric task-progress understanding.

\begingroup
\small
\linespread{1.00}\selectfont
\setlength{\itemsep}{0pt}
\setlength{\parskip}{0pt}
\setlength{\topsep}{0pt}
\setlength{\partopsep}{0pt}
\bibliographystyle{IEEEtran}
\bibliography{references}
\endgroup

\end{document}